\documentclass{article}

\usepackage{neurips_2024}

\usepackage[utf8]{inputenc} % allow utf-8 input
\usepackage[T1]{fontenc}    % use 8-bit T1 fonts
\usepackage{hyperref}       % hyperlinks
\usepackage{url}            % simple URL typesetting
\usepackage{booktabs}       % professional-quality tables
\usepackage{amsfonts}       % blackboard math symbols
\usepackage{nicefrac}       % compact symbols for 1/2, etc.
\usepackage{microtype}      % microtypography
\usepackage{xcolor}         % colors

\usepackage{times}
\usepackage{latexsym}
\usepackage{amsmath}
\usepackage{amssymb}
\usepackage{color}
\usepackage{graphicx}
\usepackage{mathtools}
\usepackage{subcaption}

\title{EWGN: Elastic Weight Generation and Context Switching in Deep Learning}

\author{%
  Shriraj P. Sawant \thanks{Use footnote for providing further information
    about author (webpage, alternative address)---\emph{not} for acknowledging
    funding agencies.} \\
  Department of Computer Science and Engineering\\
  Indian Institute of Technology Gandhinagar\\
   \\
  \texttt{sawant_shriraj@iitgn.ac.in} \\
  \And
  Krishna P. Miyapuram\thanks{Use footnote for providing further information
    about author (webpage, alternative address)---\emph{not} for acknowledging
    funding agencies.} \\
  Department of Cognitive Science\\
  Indian Institute of Technology Gandhinagar\\
   \\
  \texttt{kprasad@iitgn.ac.in} \\
}

\begin{document}

\maketitle

\begin{abstract}
  The ability to learn and retain a wide variety of tasks is a hallmark of human intelligence that has inspired research in artificial general intelligence. Continual learning approaches provide a significant step towards achieving this goal. It has been known that task variability and context switching are challenging for learning in neural networks. Catastrophic forgetting refers to the poor performance on retention of a previously learned task when a new task is being learned. Switching between different task contexts can be a useful approach to mitigate the same by preventing the interference between the varying task weights of the network. This paper introduces Elastic Weight Generative Networks (EWGN) as an idea for context switching between two different tasks. The proposed EWGN architecture uses an additional network that generates the weights of the primary network dynamically while consolidating the weights learned. The weight generation is input-dependent and thus enables context switching. Using standard computer vision datasets, namely MNIST and fashion-MNIST, we analyse the retention of previously learned task representations in Fully Connected Networks, Convolutional Neural Networks, and EWGN architectures with Stochastic Gradient Descent and Elastic Weight Consolidation learning algorithms. Understanding dynamic weight generation and context-switching ability can be useful in enabling continual learning for improved performance.
\end{abstract}

\section{Introduction}

Humans and other animals seem to be able to learn continuously, in stark contrast to artificial neural networks \cite{cichon2015branch}. For agents to become artificially intelligent, they must possess the ability to learn and retain a wide variety of tasks \cite{legg2007universal}. The propensity for knowledge of previously learned task(s), say task A is abruptly lost while information relevant to the current task, say task B is assimilated. Catastrophic forgetting is the term used to describe this phenomenon \cite{mccloskey1989catastrophic} \cite{mcclelland1995there}. This phenomenon is specifically caused when the network is trained sequentially on multiple tasks because weights that are important for task A are modified to meet task B's objectives. Therefore, it is imperative that intelligent agents exhibit continuous learning that is, the capacity to pick up new tasks without losing the ability to execute tasks that have already been trained. 

Current deep learning algorithms are rigid and static once trained and can’t adapt to new data when deployed for inferencing. In real-world scenarios, the incoming data distribution may not be static, and the trained models fail to adapt the same, especially in domains like medical imaging. One of the solutions for this was online learning, i.e., the model is retrained continuously as and when newer data are available, but this made the models forget the previously learned data.

Traditionally, methods have been developed to guarantee simultaneous availability of data from all jobs throughout training. Forgetting is prevented during learning by interleaving data from several tasks, as this allows the network's weights to be jointly tuned for optimal performance across all tasks. Deep learning approaches have been applied in this regime, which is also called the multitask learning paradigm, to train single agents to play numerous Atari games successfully \cite{mnih2015human}. Multitask learning can only be applied when tasks are given sequentially and the data are replayed to the network during training by an episodic memory system. This method (commonly referred to as system-level consolidation \cite{mcclelland1995there}) is not feasible for learning a high number of tasks since it would need a memory storage and replay ratio proportionate to the task count. 

Even though continual learning is a solution for adapting networks towards dynamic distributions, in practice it presents unique challenges for artificial neural networks. Usually, the sequential training methodology and the features anticipated from the solution characterize the challenge of continuous learning. The constant learning setting specifically focuses on non-stationary or dynamic surroundings, frequently separated into a collection of tasks that must be done sequentially, in contrast to the typical machine learning setting with a static dataset or environment. This environment might include different task transitions (smooth or discrete), different task types (unsupervised, supervised, or reinforcement learning), different task lengths and repetitions, or perhaps no clearly specified tasks at all In contrast to curriculum-based learning the learner has no say in the sequencing of tasks. In real-world scenarios, this is especially challenging because activities might not be clearly labelled, task goals and labels might flip around depending on context, and no single task might repeat over extended periods of time. 

One of the greatest obstacles to the creation of artificial general intelligence is the need to develop algorithms that facilitate ongoing learning in a dynamic way. Neuroscientific research indicates that by storing previously learned information in neocortical circuits, the mammalian brain may be able to prevent catastrophic forgetting \cite{cichon2015branch} \cite{hayashi2015labelling}. A fraction of excitatory synapses are reinforced in mice who learn new skills; this is reflected in an increase in the volume of each dendritic spine in a single neuron \cite{yang2009stably}. Crucially, these larger dendritic spines endure even after learning new tasks, which explains why performance is retained even months later \cite{yang2009stably}. The relevant skill is lost when these spines are "erased" on purpose. \cite{hayashi2015labelling} \cite{cichon2015branch}. This shows a causal relationship between task performance retention and the brain processes defending these reinforced synapses. Combining these experimental results with neurobiological models \cite{benna2016computational} suggests that task-specific synaptic consolidation, which is the process by which knowledge about how to perform a previously learned task is durably encoded in a portion of synapses that are rendered less plastic and therefore stable over long timescales, is necessary for continuous learning in the mammalian neocortex.

Little progress has been achieved in attaining complete continual learning, despite recent improvements in machine learning, particularly in deep neural networks, yielding substantial gains in performance across a range of areas \cite{10341211} \cite{lecun2015deep}. The EWC algorithm \cite{Kirkpatrick_2017} for overcoming the same is showed to have limitations and not to be commutative for given tasks \cite{10.1145/3570991.3571013}. We analyze the catastrophic forgetting phenomenon as it is the primary reason for the limitation of the model toward generalization in varying distributions. To solve this problem, we propose that learning to switch contexts between different weights might overcome catastrophic forgetting. We experimented with an instance of this context-switching network called the Elastic Weight Generation Network (EWGN) over various machine learning problems. 

\section{Methods}
\subsection{Task and Contexts}
To analyse context switching in neural networks we first need to define what is a context and task for the network. A task can be as simple as learning a single sample of the respective input-output x->y pair to as complex as simultaneously learning multiple datasets of varying distributions. So a task can be defined at each level of resolution of given data such as Distribution level, Dataset level, Class/Cluster level, and Sample level depending on the learning goal of the model. 

Whereas the context for the given learning network is the function at hand it's trying to approximate. For a given function/context same input will give the same output. Therefore weights of the network are its context. For the given task the network learns the corresponding context. Hence we define context switching in neural networks as the process of switching between the learned optimal weights of the tasks the network is trying to learn. Learning to switch, especially without task label is non trivial and what network learns, how it represents and what it retains depends on the order of observations (learning order) \cite{10.1145/3570991.3571013}. 

\begin{figure}
\begin{subfigure}{.5\textwidth}
  \centering
  \includegraphics[width=1.1\linewidth]{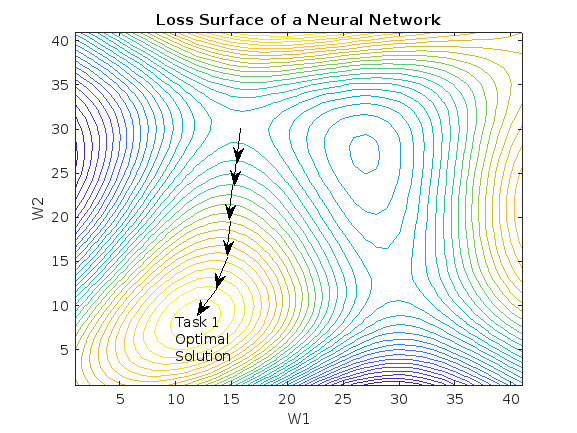}
  \caption{single set of optimal weights for all tasks}
  \label{fig:singleopt}
\end{subfigure}%
\begin{subfigure}{.5\textwidth}
  \centering
  \includegraphics[width=1.1\linewidth]{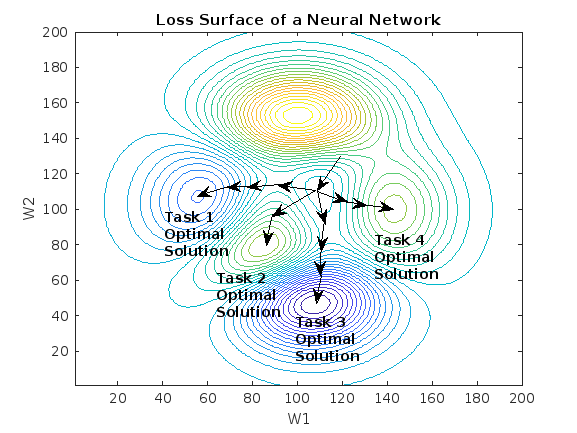}
  \caption{multiple set of optimal weights for each task}
  \label{fig:multiopt}
\end{subfigure}
\caption{loss surface of an arbitrary neural network}
\label{fig:figloss}
\end{figure}

Current neural networks learn a single set of optimal weights \ref{fig:singleopt} for single as well as multiple tasks. This might not give optimal solution as weights trying to represent multiple tasks might interfere with each other leading to catastrophic forgetting \cite{hadsell2020embracing}. One way to eliminate such interference is to freeze the weights of previous tasks or slow down their change \cite{Kirkpatrick_2017}. But the competition between the weights to represent the interfering tasks still remains. To eliminate this problem we suggest context switching between different optima of corresponding individual tasks instead of learning a single optimal weight for all the tasks thus removing the direct interference between the weights of different tasks \ref{fig:multiopt}. We posit context switching between different tasks should reflect task wise clustering of corresponding weights in the representation space of the network.

\subsection{Elastic Weight Consolidation}
Synaptic consolidation in the brain decreases the plasticity of synapses that are essential for previously performed activities, allowing for continuous learning. Elastic Weight Consolidation (EWC) is a learning algorithm that uses artificial neural networks to carry out a comparable function by restricting key parameters to remain around their initial levels. Multiple layers of linear projection are followed by element-wise non-linearities in a deep neural network. Optimizing performance during task learning involves modifying the linear projections' set of weights and biases. Various arrangements of weights will provide identical outcomes \cite{SUSSMANN1992589}; this has significance for EWC: Due to over-parameterization, there is a good chance that subsequent task B's solution, is like task A's previously discovered solution. Therefore, EWC constrains the parameters to stay in a zone of low error for task A, protecting the performance in task A while learning task B. The word "elastic" comes from the fact that this constraint is applied as a quadratic penalty, which makes it possible to visualize it as a spring that holds the parameters of the earlier solution. Crucially, this spring's stiffness should vary depending on which factors are most important to task A performance. In other words, it should be stiffer for those parameters. \cite{Kirkpatrick_2017}.

\subsection{Datasets and Training Procedure}

\begin{figure}
\begin{subfigure}{.5\textwidth}
  \centering
  \includegraphics[width=1\linewidth]{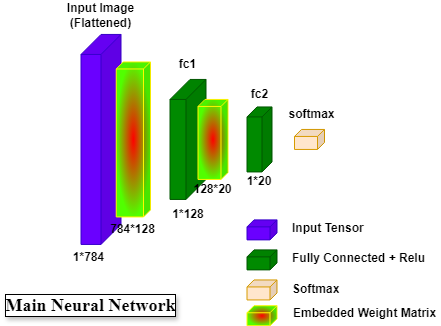}
  \caption{Fully connected MLP}
    \label{fig:mlp}
\end{subfigure}%
\begin{subfigure}{.5\textwidth}
  \centering
  \includegraphics[width=1\linewidth]{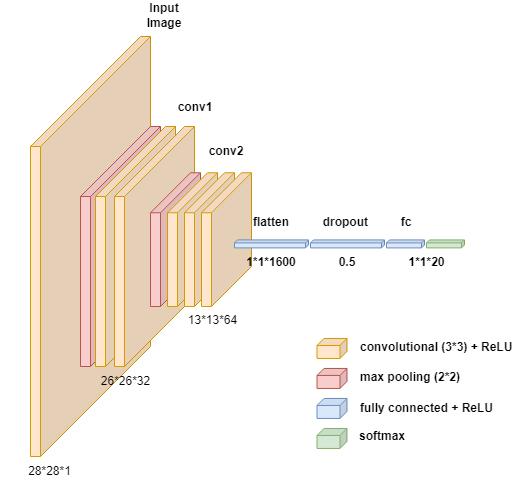}
  \caption{Convolutional Neural Network}
    \label{fig:cnn}
\end{subfigure}
\caption{Neural Network Models under analysis}
\label{fig:figmod}
\end{figure}

For our experiments we are using 2 standard computer vision datasets namely MNIST \cite{lecun-mnisthandwrittendigit-2010} and Fashion-MNIST \cite{xiao2017fashionmnist} as 2 different tasks to be learned sequentially by the corresponding neural networks. We will be analysing 3 different types of neural network architectures, a fully connected multi layer perceptron (MLP) \ref{fig:mlp}, a convolutional neural network (CNN) \ref{fig:cnn} and our own proposed elastic weight generation network (EWGN) \ref{fig:ewgn}. We will also be testing both the Stochastic Gradient Descent (SGD) \cite{ruder2016overview}  and the Elastic Weight Consolidation (EWC) \cite{Kirkpatrick_2017} learning algorithms for the same. The corresponding network parameters are given in the respective figures. 

For training all the networks will be optimised using Adam variant of the SGD \cite{kingma2017adam} for 10 epochs of each task with learning rate varying from 0.001 to 0.005. Batch size is limited to 1 sample per step. Each respective datasets is split into 60000 images for training and 10000 images for testing.  Both of the tasks are trained in an extended label fashion i.e. models are provided with total no. of classes as labelled vector of both tasks (i.e. 10 + 10 = 20 dimensional one hot label in this case). We will be analysing the models in both possible orders i.e. Task A first then Task B and vice versa for understanding the implication of learning order of the tasks.

\subsection{Proposed Model for Context Switching}

\begin{figure}[!htb]
    \centering
    \includegraphics[scale=0.5]{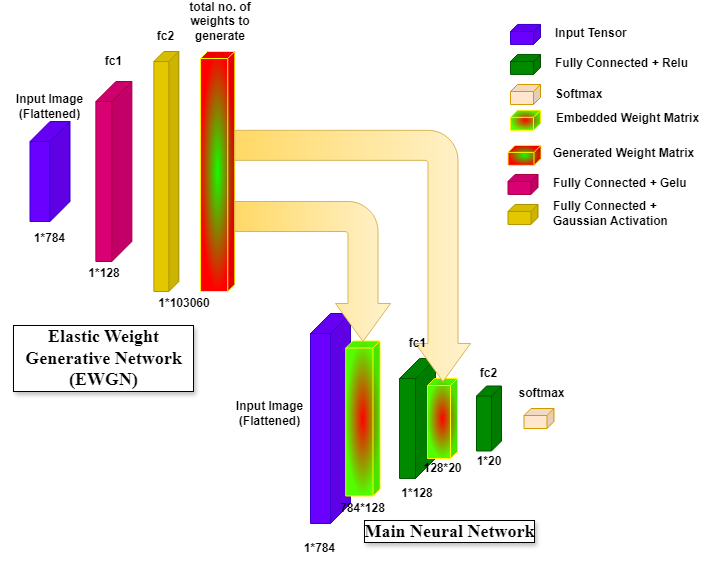}
    \caption{The Architecture of an Artificial Neural Network augmented by the context switching Elastic weight generative network (EWGN)}
    \label{fig:ewgn}
\end{figure}

We implement an instance of the proposed idea of the hypothetical context switching network, called the Elastic Weight Generation Network (EWGN) \ref{fig:ewgn} via augmenting a main fully connected neural network having same architecture as the MLP described previously \ref{fig:mlp}. The weight generative network is inspired by Hypernetworks \cite{chauhan2023brief} \cite{ha2016hypernetworks} \cite{hemati2023partial}. The Elastic in the EWGN stands for the EWC algorithm used to consolidate the weights of the EWGN. Here the weights of the main network are adapted by the EWGN during both training and inferencing phases thus making the complete neural network adaptive while testing as well. We hypothesize that such an augmented neural network with EWGN is a Universal “multi” function approximator \ref{fig:ewgn}.

The EWGN model learns to context switch in the input dependent fashion.   First the EWGN takes the given input tensor, and generates the corresponding no. of weights for the main network. This weights are then embedded into the main network at each step of training of individual sample. The main network then takes the input tensor and predicts the output with the weights generated by the EWGN. The loss thus calculated for this predictions is then fed back to train the weights of the EWGN. Here the main network is not backpropagated and the task label is not provided and the model has to inherently learn to switch between the tasks by generating corresponding weights.

The conceptual idea behind EWC was to dampen the adaptation of weights important to the previously learned tasks. A form of regularization is achieved via the usage of the Fisher information matrix. But that doesn’t solve the core problem of catastrophic forgetting, which is the tug of war between the weights of the network to represent the competing sequentially learned tasks. It finds a middle point between the optima of both individual tasks which is lesser than individual optima. This problem arises because the network is trying to learn a single set of weights. To overcome the same, the idea behind EWGN is to learn multiple sets of weights instead of one for all the given tasks and context switch between them and transform the goal of learning single set of optimal weights for all the task into generating multiple optimal set of weights corresponding to each of the task. This should eliminate the competition between the weights and perhaps we can completely bypass the problem of catastrophic forgetting.

\subsection{Evaluation Metrics}

For evaluating the said models we will be using standard accuracy metric and retention rate. Each of the model will be subsequently trained with Task A and Task B and their training and test accuracy's in sequence is documented in corresponding columns of the table \ref{tab:my-table}. For calculating the retention rate we subtract the difference between test accuracy's of first task A before and after learning second task B from 100. This metrics are chosen to analyse the catastrophic forgetting of previously learned task. Further we calculate the classwise confusion matrices and the UMAP \cite{mcinnes2020umap} projections of generated weights to analyse the dynamic weight generation and context switching ability of the model.

\section{Experiments and Results}
\subsection{Setup and Compute Resources}
For implementing the neural networks we used tensorflow 2.0+ \cite{tensorflow2015-whitepaper} neural network library on CUDA 11.4 \cite{cuda} with driver version 470.182. Nvidia Quadro 5000 GPU was used for training our models. The experiment used 16 GB of VRAM to train the models. But the GPU is shared across our lab for multiple experiments and hence both time and memory limited. For EWGN we had to use Eager Execution mode of the tensorflow library to enable changing of weights of the main network by EWGN at every step. Eager mode is very slow compared to graph execution mode of tensorflow and takes more time to train EWGN, roughly 24 hours full 10 epochs on both the datasets.

\subsection{Model Performance Analysis}

Here we describe the performance and the retention rate of the corresponding models \ref{tab:my-table}. Each of the columns starting from Task A train accuracy to the second last column of the table i.e. Task A test accuracy are in the sequence of the corresponding order of the training and testing of Task A and Task B. Here we observe that EWGN-MLP model has highest retention rate of the previously learned Task A in both learning orders i.e from MNIST to FMNIST and vice versa.

% \usepackage{booktabs}
% \usepackage{graphicx}
\begin{table}[]
\centering
\caption{Performance and Retention metrics for sequential training}
\label{tab:my-table}
\resizebox{\textwidth}{!}{%
\begin{tabular}{@{}lllllllllll@{}}
\toprule
\textbf{Model} &
  \textbf{\begin{tabular}[c]{@{}l@{}}Learning \\ Algorithm\end{tabular}} &
  \textbf{Task A} &
  \textbf{Task B} &
  \textbf{\begin{tabular}[c]{@{}l@{}}Task A \\ Train Acc.\end{tabular}} &
  \textbf{\begin{tabular}[c]{@{}l@{}}Task A \\ Test Acc.\end{tabular}} &
  \textbf{\begin{tabular}[c]{@{}l@{}}Task B \\ Test Acc.\end{tabular}} &
  \textbf{\begin{tabular}[c]{@{}l@{}}Task B \\ Train Acc.\end{tabular}} &
  \textbf{\begin{tabular}[c]{@{}l@{}}Task B \\ Test Acc.\end{tabular}} &
  \textbf{\begin{tabular}[c]{@{}l@{}}Task A \\ Test Acc.\end{tabular}} &
  \textbf{Retention} \\ \midrule
MLP      & SGD & MNIST  & FMNIST & 99.20\% & 97.64\% & 10.58\% & 90.30\% & 87.85\% & 29.77\% & 32.13\%          \\
MLP      & EWC & MNIST  & FMNIST & N/A     & N/A     & N/A     & 87.11\% & 85.47\% & 95.38\% & 97.74\%          \\
MLP      & SGD & FMNIST & MNIST  & 90.30\% & 87.94\% & 8.26\%  & 99.32\% & 97.65\% & 22.07\% & 34.13\%          \\
MLP      & EWC & FMNIST & MNIST  & N/A     & N/A     & N/A     & 94.27\% & 94.05\% & 35.20\% & 47.26\%          \\
CNN      & SGD & MNIST  & FMNIST & 98.71\% & 99.06\% & 12.56\% & 89.93\% & 90.12\% & 15.93\% & 16.87\%          \\
CNN      & EWC & MNIST  & FMNIST & N/A     & N/A     & N/A     & 77.31\% & 81.84\% & 83.51\% & 84.45\%          \\
CNN      & SGD & FMNIST & MNIST  & 89.69\% & 90.17\% & 12.37\% & 98.64\% & 99.02\% & 22.69\% & 32.52\%          \\
CNN      & EWC & FMNIST & MNIST  & N/A     & N/A     & N/A     & 93.32\% & 96.96\% & 46.64\% & 56.47\%          \\
WGN-MLP  & SGD & MNIST  & FMNIST & 96.45\% & 95.74\% & 0\%     & 84.14\% & 83.15\% & 25.05\% & 29.31\%          \\
EWGN-MLP & EWC & MNIST  & FMNIST & 98.95\% & 97.80\% & 0\%     & 75.06\% & 73.93\% & 97.67\% & \textbf{99.87\%} \\
WGN-MLP  & SGD & FMNIST & MNIST  & 89.58\% & 87.19\% & 0\%     & 98.90\% & 94.66\% & 28.72\% & 41.53\%          \\
EWGN-MLP & EWC & FMNIST & MNIST  & 89.5\%  & 87.19\% & 0\%     & 62.44\% & 63.35\% & 68.46\% & \textbf{81.27\%} \\ \bottomrule
\end{tabular}%
}
\end{table}

\subsection{Representation Analysis}
\begin{figure}
\begin{subfigure}{.5\textwidth}
  \centering
    \includegraphics[width=.8\linewidth]{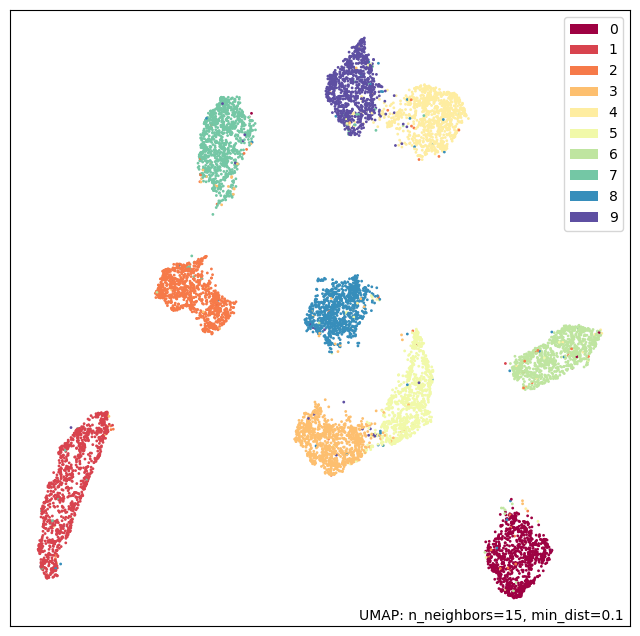}
    \caption{Generated weights for single task (MNIST) dataset}
    \label{fig:mpred}
\end{subfigure}%
\begin{subfigure}{.5\textwidth}
  \centering
  \includegraphics[width=.8\linewidth]{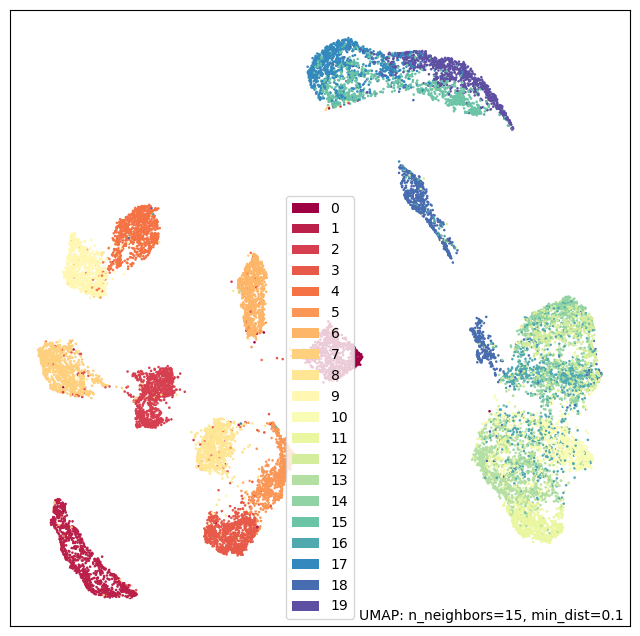}
  \caption{Retained weights of both tasks in EWGN}
    \label{fig:mfseqwt}
\end{subfigure}
\begin{subfigure}{.5\textwidth}
  \centering
  \includegraphics[width=.8\linewidth]{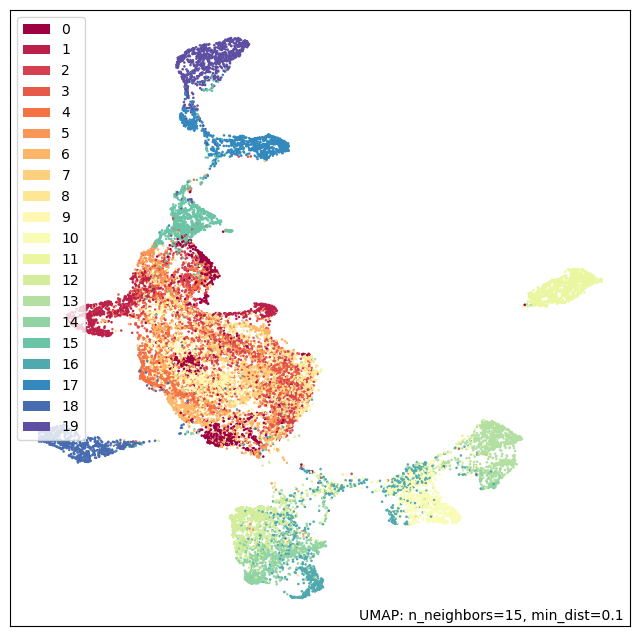}
  \caption{Forgotten weights of MNIST task without EWGN}
    \label{fig:mforgot}
\end{subfigure}
\begin{subfigure}{.5\textwidth}
  \centering
  \includegraphics[width=.8\linewidth]{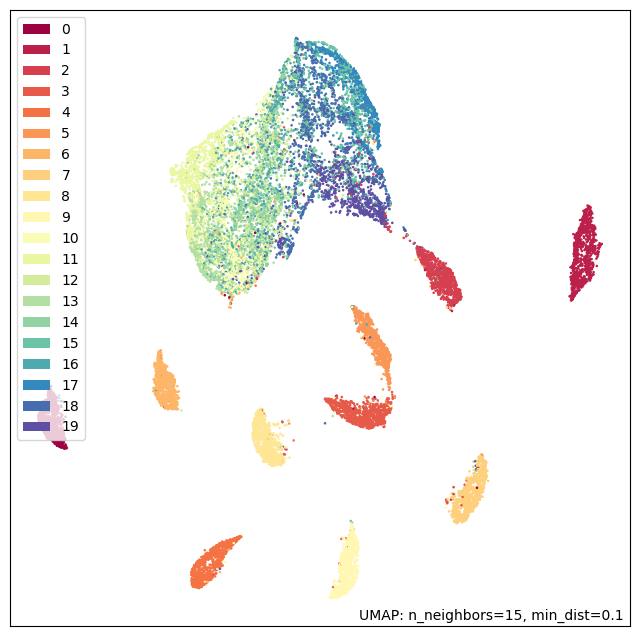}
  \caption{Forgotten weights of fMNIST task without EWGN}
    \label{fig:fmforgot}
\end{subfigure}
\caption{Uniform Manifold Approximations and Projections}
\label{fig:figumap}
\end{figure}

\begin{figure}
\begin{subfigure}{.5\textwidth}
  \centering
    \includegraphics[width=.8\linewidth]{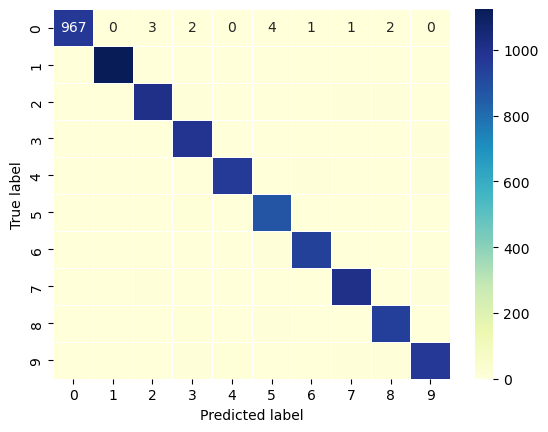}
    \caption{EWGN performance on single task (MNIST)}
    \label{fig:mpredcm}
\end{subfigure}%
\begin{subfigure}{.5\textwidth}
  \centering
  \includegraphics[width=.8\linewidth]{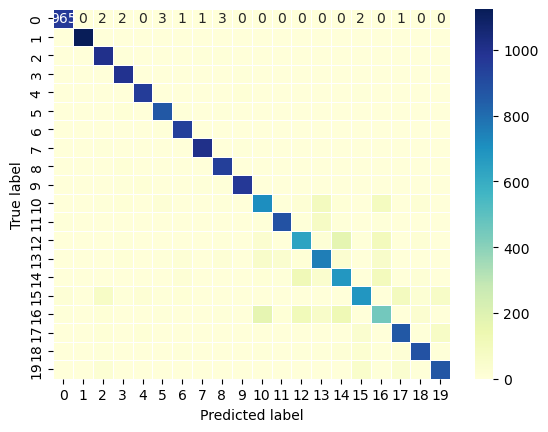}
  \caption{EWGN performance of both tasks in sequence}
    \label{fig:mnfcm}
\end{subfigure}
\begin{subfigure}{.5\textwidth}
  \centering
  \includegraphics[width=.8\linewidth]{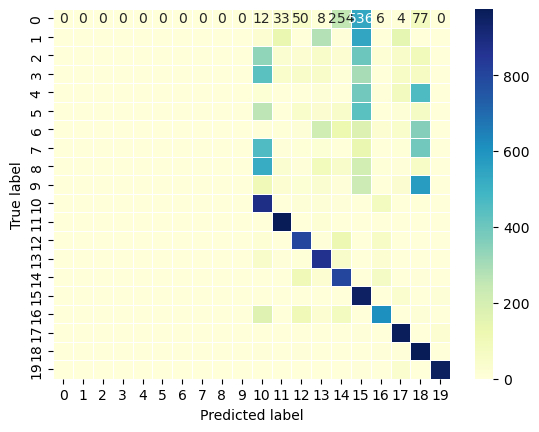}
  \caption{Forgotten accuracy of MNIST task without EWGN}
    \label{fig:mforgotcm}
\end{subfigure}
\begin{subfigure}{.5\textwidth}
  \centering
  \includegraphics[width=.8\linewidth]{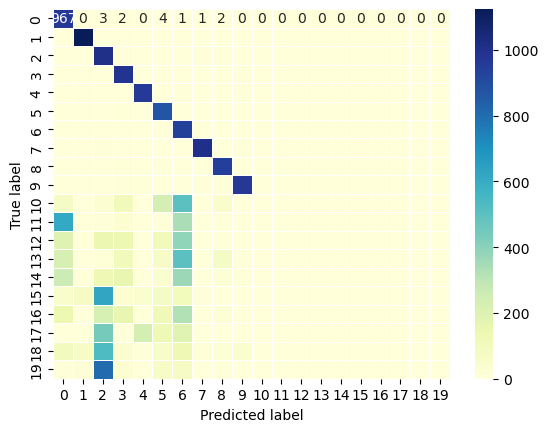}
  \caption{Forgotten accuracy of fMNIST task without EWGN}
    \label{fig:fmforgotcm}
\end{subfigure}
\caption{Confusion Matrices}
\label{fig:figcm}
\end{figure}

For each of the tasks we also calculated the Uniform Manifold Approximations and Projections (UMAP) \cite{mcinnes2020umap} of the generated weights along with the confusion matrices of predictions made by the EWGN and non EWGN networks for understanding the context switching capability. The figures \ref{fig:mpred} and \ref{fig:mpredcm} shows the UMAP projections of the generated weights for first task A of MNIST dataset and its corresponding predictions in confusion matrix. \ref{fig:mfseqwt} and \ref{fig:mnfcm} shows the generated weights and corresponding predictions after EWGN is trained with the second task B of fMNIST after MNIST. Here we can see that EWGN is able to retain the clusters of previously learned task weights while learning the new one.

The rest of the figures \ref{fig:mforgot}, \ref{fig:fmforgot} and their corresponding predictions \ref{fig:mforgotcm} and \ref{fig:fmforgotcm} shows the case where previously learned task weights aren't retained, evident from the dispersion of corresponding clusters and the prediction performance of the previously learned tasks.

\section{Discussion}
Initially we weren't able to merge EWC learning algorithm with WGN architecture, the model simply didn't worked before. The core problem was the fisher information matrix of EWC and the initial tanh output activation of WGN. Output of WGN is not positive logits unlike prediction models. Its the weights of the network. applying log likelihood to that gave lot of NaN values and zeroes owing to -ve weights and also numerical unstability. That infact caused the failure of consolidation of previous tasks. 

We found 2 possibilities that could solve this problem, one was to clip the -ve predictions of weights before feeding it to fisher matrix computation. But that would lead -ve tanh outputs to be useless. 
So second way was to limit the output of WGN to positive only. Using sigmoid was the easiest choice, but using only positive output with monotonic function would limit the universal function approximation ability of the network. 

So third way was to use a non monotonic positive function. Many functions could be used. Studies on distribution of neural network weights revealed that gaussian distribution weighted networks doesn't only reduce overfitting due to large weights but also helps in generalisability of the said network \cite{gawlikowski2023survey}. So we used a custom gaussian activation function for the WGN and also used epsilon clipping for the fisher information matrix to handle the edge cases where the weights are either 0 or 1 which can make log likelihood computation unstable. The gaussian activation function also inherently normalises the generated weights of the network without layernorm \cite{ba2016layer} or batchnorms \cite{ioffe2015batch}. Experimenting with different types of activation's and distribution of learned weights in known networks might help in improving the generated weights of the network.

The UMAP projections of the generated task weights \ref{fig:mfseqwt} in EWGN shows that generated weights are clustering around the respective tasks and also around the classes of the individual tasks suggesting existence of multiple optimal weights not just for an individual task but also individual classes, confirming our original definition of resolution of tasks at each level of datasets, classes and samples. Despite forgetting of previously learned tasks, the generated weights in \ref{fig:mforgot} \ref{fig:fmforgot} still suggest possibility of determining the different tasks if not the classes as task level cluster of weights is still present.

What causes forgetting is the competition between weights and eliminating that is the next step to deeper learning. To eliminate this competition of weights we must make the existing deep neural networks adaptive and dynamic to varying distributions of the data. We need to capture and control the hyperplanes/decision boundaries of a Neural Network for learning nonstationary and dynamic distributions. To achieve that, first, we need to stabilize and limit the plasticity of weights during the training phase to prevent catastrophic forgetting. Secondly, we need to induce controlled plasticity in the weights of the network during the inference phase for handling nonstationarity. The key for that is to make the hyperplanes of the neural networks not fixed during inference and continuously adapt according to a context-switching network that models the distribution of weights of the base network. Our primitive implementation of such a context switching network, called the elastic weight generation network, tries to achieve some of these goals and the primary results give us hope that this might be the right path towards solving this problem.

\section{Limitations}
Although our experiments studied catastrophic forgetting phenomena in multiple architectures and analysed the idea of context switching via weight generation, the study is limited to standard datasets and might not reflect the real world results. The assumption behind standard datasets is the IID (Independent and Identical Distribution) which might not hold for practical cases in real world. The performance of the model and retention rate might not hold for tasks with different distributions than MNIST and FMNIST datasets and for more than 2 tasks. 

Also the relatively small size of the networks might affect the model retention rate which needs further testing. Due to hardware limitations we weren't able to experiment with bigger networks with larger iterations. The eager execution mode of tensorflow needed for EWGN model slows down the training compared to the standard models which runs on fast graph execution mode. Further research is needed to implement EWGN model in graph execution mode to speed up model training. Most importantly the problem of task order drastically affecting the retention capability of the said neural networks needs further research\cite{10.1145/3570991.3571013}. Having dynamic consolidation strength for varying task orders might help in mitigating the same \cite{benzing2021unifying}.

\section{Broader Impacts}
This research can influence the broader practices within AI and machine learning by encouraging the adoption of multi-weight strategies and context-aware training methodologies. Addressing the critical issue of catastrophic forgetting, will pave the way for more resilient, adaptable, and intelligent neural network models. Such approaches can be integrated into existing machine learning frameworks, leading to more robust and flexible AI models \cite{wang2024comprehensive}. Systems that remember previous tasks and decisions can offer better transparency and traceability, important for accountability in critical applications like healthcare, finance, and law enforcement. Additionally, the insights gained from this research can inform the development of new algorithms and techniques for other types of neural networks, further advancing the state of the art. The broader impact of this research lies in its potential to significantly advance the field of continual learning \cite{hadsell2020embracing}, improve practical applications across various domains, and contribute to the ethical and fair deployment of AI systems. 

\section{Conclusion}

Sequential and nonstationary Learning exposes the limitations of neural networks. A neural network fails to retain previously learned tasks due to Catastrophic forgetting. The primary reason behind this is the tug of war between the weights of the networks when there is a conflict between representations of two different tasks. Our hypothesis was to eliminate this competition between weights via context switching to a different set of weights for the respective tasks. The elastic weight-generating network [EWGN] transforms the sequential learning problem of 2 different tasks into a common problem of optimal weight generation. Via context switching between the weights it eliminates the direct competition between the weights of the main network for different tasks. The context-switching and dynamic adaptation ability for neural networks might be the key to pave the way towards artificial general intelligence.

\bibliography{neurips_2024}

\begin{thebibliography}{10}

\bibitem{tensorflow2015-whitepaper}
Mart\'{\i}n Abadi, Ashish Agarwal, Paul Barham, Eugene Brevdo, Zhifeng Chen, Craig Citro, Greg~S. Corrado, Andy Davis, Jeffrey Dean, Matthieu Devin, Sanjay Ghemawat, Ian Goodfellow, Andrew Harp, Geoffrey Irving, Michael Isard, Yangqing Jia, Rafal Jozefowicz, Lukasz Kaiser, Manjunath Kudlur, Josh Levenberg, Dan Man\'{e}, Rajat Monga, Sherry Moore, Derek Murray, Chris Olah, Mike Schuster, Jonathon Shlens, Benoit Steiner, Ilya Sutskever, Kunal Talwar, Paul Tucker, Vincent Vanhoucke, Vijay Vasudevan, Fernanda Vi\'{e}gas, Oriol Vinyals, Pete Warden, Martin Wattenberg, Martin Wicke, Yuan Yu, and Xiaoqiang Zheng.
\newblock {TensorFlow}: Large-scale machine learning on heterogeneous systems, 2015.
\newblock Software available from tensorflow.org.

\bibitem{ba2016layer}
Jimmy~Lei Ba, Jamie~Ryan Kiros, and Geoffrey~E. Hinton.
\newblock Layer normalization, 2016.

\bibitem{benna2016computational}
Marcus~K Benna and Stefano Fusi.
\newblock Computational principles of synaptic memory consolidation.
\newblock {\em Nature neuroscience}, 19(12):1697--1706, 2016.

\bibitem{benzing2021unifying}
Frederik Benzing.
\newblock Unifying regularisation methods for continual learning, 2021.

\bibitem{chauhan2023brief}
Vinod~Kumar Chauhan, Jiandong Zhou, Ping Lu, Soheila Molaei, and David~A. Clifton.
\newblock A brief review of hypernetworks in deep learning, 2023.

\bibitem{cichon2015branch}
Joseph Cichon and Wen-Biao Gan.
\newblock Branch-specific dendritic ca2+ spikes cause persistent synaptic plasticity.
\newblock {\em Nature}, 520(7546):180--185, 2015.

\bibitem{gawlikowski2023survey}
Jakob Gawlikowski, Cedrique Rovile~Njieutcheu Tassi, Mohsin Ali, Jongseok Lee, Matthias Humt, Jianxiang Feng, Anna Kruspe, Rudolph Triebel, Peter Jung, Ribana Roscher, et~al.
\newblock A survey of uncertainty in deep neural networks.
\newblock {\em Artificial Intelligence Review}, 56(Suppl 1):1513--1589, 2023.

\bibitem{ha2016hypernetworks}
David Ha, Andrew Dai, and Quoc~V. Le.
\newblock Hypernetworks, 2016.

\bibitem{hadsell2020embracing}
Raia Hadsell, Dushyant Rao, Andrei~A Rusu, and Razvan Pascanu.
\newblock Embracing change: Continual learning in deep neural networks.
\newblock {\em Trends in cognitive sciences}, 24(12):1028--1040, 2020.

\bibitem{hayashi2015labelling}
Akiko Hayashi-Takagi, Sho Yagishita, Mayumi Nakamura, Fukutoshi Shirai, Yi~I Wu, Amanda~L Loshbaugh, Brian Kuhlman, Klaus~M Hahn, and Haruo Kasai.
\newblock Labelling and optical erasure of synaptic memory traces in the motor cortex.
\newblock {\em Nature}, 525(7569):333--338, 2015.

\bibitem{hemati2023partial}
Hamed Hemati, Vincenzo Lomonaco, Davide Bacciu, and Damian Borth.
\newblock Partial hypernetworks for continual learning, 2023.

\bibitem{ioffe2015batch}
Sergey Ioffe and Christian Szegedy.
\newblock Batch normalization: Accelerating deep network training by reducing internal covariate shift, 2015.

\bibitem{kingma2017adam}
Diederik~P. Kingma and Jimmy Ba.
\newblock Adam: A method for stochastic optimization, 2017.

\bibitem{Kirkpatrick_2017}
James Kirkpatrick, Razvan Pascanu, Neil Rabinowitz, Joel Veness, Guillaume Desjardins, Andrei~A. Rusu, Kieran Milan, John Quan, Tiago Ramalho, Agnieszka Grabska-Barwinska, Demis Hassabis, Claudia Clopath, Dharshan Kumaran, and Raia Hadsell.
\newblock Overcoming catastrophic forgetting in neural networks.
\newblock {\em Proceedings of the National Academy of Sciences}, 114(13):3521–3526, March 2017.

\bibitem{lecun2015deep}
Yann LeCun, Yoshua Bengio, and Geoffrey Hinton.
\newblock Deep learning.
\newblock {\em nature}, 521(7553):436--444, 2015.

\bibitem{lecun-mnisthandwrittendigit-2010}
Yann LeCun and Corinna Cortes.
\newblock {MNIST} handwritten digit database.
\newblock 2010.

\bibitem{legg2007universal}
Shane Legg and Marcus Hutter.
\newblock Universal intelligence: A definition of machine intelligence, 2007.

\bibitem{mcclelland1995there}
James~L McClelland, Bruce~L McNaughton, and Randall~C O'Reilly.
\newblock Why there are complementary learning systems in the hippocampus and neocortex: insights from the successes and failures of connectionist models of learning and memory.
\newblock {\em Psychological review}, 102(3):419, 1995.

\bibitem{mccloskey1989catastrophic}
Michael McCloskey and Neal~J Cohen.
\newblock Catastrophic interference in connectionist networks: The sequential learning problem.
\newblock In {\em Psychology of learning and motivation}, volume~24, pages 109--165. Elsevier, 1989.

\bibitem{mcinnes2020umap}
Leland McInnes, John Healy, and James Melville.
\newblock Umap: Uniform manifold approximation and projection for dimension reduction, 2020.

\bibitem{mnih2015human}
Volodymyr Mnih, Koray Kavukcuoglu, David Silver, Andrei~A Rusu, Joel Veness, Marc~G Bellemare, Alex Graves, Martin Riedmiller, Andreas~K Fidjeland, Georg Ostrovski, et~al.
\newblock Human-level control through deep reinforcement learning.
\newblock {\em nature}, 518(7540):529--533, 2015.

\bibitem{cuda}
NVIDIA, Péter Vingelmann, and Frank~H.P. Fitzek.
\newblock Cuda, release: 10.2.89, 2020.

\bibitem{ruder2016overview}
Sebastian Ruder.
\newblock An overview of gradient descent optimization algorithms.
\newblock {\em arXiv preprint arXiv:1609.04747}, 2016.

\bibitem{10.1145/3570991.3571013}
Shriraj~Pramod Sawant.
\newblock Understanding catastrophic forgetting for adaptive deep learning.
\newblock In {\em Proceedings of the 6th Joint International Conference on Data Science \& Management of Data (10th ACM IKDD CODS and 28th COMAD)}, CODS-COMAD '23, page 282–283, New York, NY, USA, 2023. Association for Computing Machinery.

\bibitem{SUSSMANN1992589}
Héctor~J. Sussmann.
\newblock Uniqueness of the weights for minimal feedforward nets with a given input-output map.
\newblock {\em Neural Networks}, 5(4):589--593, 1992.

\bibitem{wang2024comprehensive}
Liyuan Wang, Xingxing Zhang, Hang Su, and Jun Zhu.
\newblock A comprehensive survey of continual learning: Theory, method and application.
\newblock {\em IEEE Transactions on Pattern Analysis and Machine Intelligence}, 2024.

\bibitem{10341211}
B.~Wickramasinghe, G.~Saha, and K.~Roy.
\newblock Continual learning: A review of techniques, challenges and future directions.
\newblock {\em IEEE Transactions on Artificial Intelligence}, 1(01):1--21, dec 5555.

\bibitem{xiao2017fashionmnist}
Han Xiao, Kashif Rasul, and Roland Vollgraf.
\newblock Fashion-mnist: a novel image dataset for benchmarking machine learning algorithms, 2017.

\bibitem{yang2009stably}
Guang Yang, Feng Pan, and Wen-Biao Gan.
\newblock Stably maintained dendritic spines are associated with lifelong memories.
\newblock {\em Nature}, 462(7275):920--924, 2009.

\end{thebibliography}
\bibliographystyle{plain}

%%%%%%%%%%%%%%%%%%%%%%%%%%%%%%%%%%%%%%%%%%%%%%%%%%%%%%%%%%%%

\appendix

\section{Appendix / supplemental material}

The corresponding code files and additional experiments are shared seperately and also available at the github repository in jupyter notebook format https://github.com/iam-sr13/ewgn.

%%%%%%%%%%%%%%%%%%%%%%%%%%%%%%%%%%%%%%%%%%%%%%%%%%%%%%%%%%%%

\newpage
\section*{NeurIPS Paper Checklist}

\begin{enumerate}

\item {\bf Claims}
    \item[] Question: Do the main claims made in the abstract and introduction accurately reflect the paper's contributions and scope?
    \item[] Answer: \answerYes{} % Replace by \answerYes{}, \answerNo{}, or \answerNA{}.
    \item[] Justification: The claims made in abstract are respectively written in the methods, results and discussion section. The 2.4 proposed model subsection defines the EWGN model of the paper. The experimental details are mentioned in the 2.3 datasets and training procedure subsection with their corresponding results in the 3.2 model performance analysis subsection. 

\item {\bf Limitations}
    \item[] Question: Does the paper discuss the limitations of the work performed by the authors?
    \item[] Answer: \answerYes{} % Replace by \answerYes{}, \answerNo{}, or \answerNA{}.
    \item[] Justification: Limitations section

\item {\bf Theory Assumptions and Proofs}
    \item[] Question: For each theoretical result, does the paper provide the full set of assumptions and a complete (and correct) proof?
    \item[] Answer: \answerNA{} % Replace by \answerYes{}, \answerNo{}, or \answerNA{}.
    \item[] Justification: NA

    \item {\bf Experimental Result Reproducibility}
    \item[] Question: Does the paper fully disclose all the information needed to reproduce the main experimental results of the paper to the extent that it affects the main claims and/or conclusions of the paper (regardless of whether the code and data are provided or not)?
    \item[] Answer: \answerYes{} % Replace by \answerYes{}, \answerNo{}, or \answerNA{}.
    \item[] Justification: Methods section and supplementary code files provided. 2.3 Datasets and Training Procedure subsection describes details for reproducing experiments. The figures of model clearly describe the archtecture parameters used for the experiments. Also full jupyter notebooks with all the experiments are provided as supplementary files.

\item {\bf Open access to data and code}
    \item[] Question: Does the paper provide open access to the data and code, with sufficient instructions to faithfully reproduce the main experimental results, as described in supplemental material?
    \item[] Answer: \answerYes{} % Replace by \answerYes{}, \answerNo{}, or \answerNA{}.
    \item[] Justification: Experiments and Code files provided as jupyter notebooks.

\item {\bf Experimental Setting/Details}
    \item[] Question: Does the paper specify all the training and test details (e.g., data splits, hyperparameters, how they were chosen, type of optimizer, etc.) necessary to understand the results?
    \item[] Answer: \answerYes{} % Replace by \answerYes{}, \answerNo{}, or \answerNA{}.
    \item[] Justification: Methods and code files. Especially the section 2.3 Dataset and Training procedure.

\item {\bf Experiment Statistical Significance}
    \item[] Question: Does the paper report error bars suitably and correctly defined or other appropriate information about the statistical significance of the experiments?
    \item[] Answer:  \answerNo{} % Replace by \answerYes{}, \answerNo{}, or \answerNA{}.
    \item[] Justification: Due to hardware limitations and slow execution of eager mode of tensorflow running multiple experiments weren't possible in time.

\item {\bf Experiments Compute Resources}
    \item[] Question: For each experiment, does the paper provide sufficient information on the computer resources (type of compute workers, memory, time of execution) needed to reproduce the experiments?
    \item[] Answer: \answerYes{} % Replace by \answerYes{}, \answerNo{}, or \answerNA{}.
    \item[] Justification: Methods and code files. The section 3.1 Setup and Compute resources mentions details regarding memory and compute time.
    
\item {\bf Code Of Ethics}
    \item[] Question: Does the research conducted in the paper conform, in every respect, with the NeurIPS Code of Ethics \url{https://neurips.cc/public/EthicsGuidelines}?
    \item[] Answer: \answerYes{} % Replace by \answerYes{}, \answerNo{}, or \answerNA{}.
    \item[] Justification: 

\item {\bf Broader Impacts}
    \item[] Question: Does the paper discuss both potential positive societal impacts and negative societal impacts of the work performed?
    \item[] Answer: \answerYes{} % Replace by \answerYes{}, \answerNo{}, or \answerNA{}.
    \item[] Justification: Broader Impacts section
    
\item {\bf Safeguards}
    \item[] Question: Does the paper describe safeguards that have been put in place for responsible release of data or models that have a high risk for misuse (e.g., pretrained language models, image generators, or scraped datasets)?
    \item[] Answer: \answerNA{} % Replace by \answerYes{}, \answerNo{}, or \answerNA{}.
    \item[] Justification: NA

\item {\bf Licenses for existing assets}
    \item[] Question: Are the creators or original owners of assets (e.g., code, data, models), used in the paper, properly credited and are the license and terms of use explicitly mentioned and properly respected?
    \item[] Answer: \answerYes{} % Replace by \answerYes{}, \answerNo{}, or \answerNA{}.
    \item[] Justification: Methods section

\item {\bf New Assets}
    \item[] Question: Are new assets introduced in the paper well documented and is the documentation provided alongside the assets?
    \item[] Answer: \answerYes{} % Replace by \answerYes{}, \answerNo{}, or \answerNA{}.
    \item[] Justification: The model figures are done by ourselves using draw.io tools and rest of the model training details are mentioned in the paper.

\item {\bf Crowdsourcing and Research with Human Subjects}
    \item[] Question: For crowdsourcing experiments and research with human subjects, does the paper include the full text of instructions given to participants and screenshots, if applicable, as well as details about compensation (if any)? 
    \item[] Answer: \answerNA{} % Replace by \answerYes{}, \answerNo{}, or \answerNA{}.
    \item[] Justification: The datasets used are publicly available and open sourced. No human subjects were used in our experiments and are not applicable.

\item {\bf Institutional Review Board (IRB) Approvals or Equivalent for Research with Human Subjects}
    \item[] Question: Does the paper describe potential risks incurred by study participants, whether such risks were disclosed to the subjects, and whether Institutional Review Board (IRB) approvals (or an equivalent approval/review based on the requirements of your country or institution) were obtained?
    \item[] Answer: \answerNA{} % Replace by \answerYes{}, \answerNo{}, or \answerNA{}.
    \item[] Justification: The datasets used are publicly available and open sourced. No human subjects were used in our experiments and are not applicable.

\end{enumerate}

\end{document}